\documentclass[journal]{IEEEtran}

\usepackage{amsmath, amsfonts, amssymb}

\usepackage{graphicx}
\usepackage{pgfplots}
\usepackage{pgfplotstable}
\usepackage{tikz}
\usepackage{tikzscale}
\pgfplotsset{compat=newest}
\usepgfplotslibrary{groupplots}

\usepackage{siunitx}      
\usepackage{array}        
\usepackage{booktabs}     
\usepackage{multirow}     
\usepackage{tabularx}     
\usepackage{longtable}    
\usepackage{makecell}     
\usepackage{threeparttable} 
\usepackage{colortbl}     

\usepackage{cite}
\usepackage{url}
\usepackage[acronym]{glossaries}
\usepackage{hyperref}
\hypersetup{
	colorlinks,
	linkcolor={black},
	citecolor={black},
	urlcolor={blue!80!black}
}

\usepackage[utf8]{inputenc}
\usepackage{lmodern}
\usepackage{csquotes}
\usepackage{placeins}     
\usepackage{flushend}     
\usepackage[font=small,labelfont=bf]{caption}
\usepackage{balance}

\usepackage{subfig} 

\usepackage{xcolor}       

\newacronym{SP}{SP}{Stewart Platform}
\newacronym{DoF}{DoF}{degree of freedom}

\newcommand{\vek}[1]{{\mathbf #1}}

\newcommand{\sv}[1]{\boldsymbol{#1}}

\newcommand{\rev}[1]{{#1}}

\usepackage{multicol}
\usepackage{fancyhdr}
\fancypagestyle{firstpage}{
	\fancyhf{}
	\fancyhead[C]{This work has been submitted to the IEEE for possible publication. Copyright may be transferred without notice, after which this version may no longer be accessible.}
}
\begin{document}
\pagenumbering{gobble} 
\thispagestyle{firstpage}
\title{\rev{An Open-Source ROS\,2 Framework} for Shared Control in Surgical Teleoperation: Integrating the Haply Inverse3 with the \\ da Vinci Research Kit}

\author{Balint~Varga,~\IEEEmembership{Member,~IEEE,} %
        Sebastian~Fritz, Gergely Kiss, Kristof~Takacs, \\ Peter~Galambos,~\IEEEmembership{Senior Member,~IEEE,}%
\thanks{This work is related to the MedLaBotX project, financed under 2024-1.2.3-HU-RIZONT-00069.}
\thanks{B. Varga and S. Fritz are with the Institute of Control Systems, Karlsruhe Institute of Technology, 76131 Karlsruhe, Germany (e-mail: balint.varga2@kit.edu).}
\thanks{G. Kiss, K. Takacs, and P. Galambos are with the Antal Bejczy Center for Intelligent Robotics, Research and Innovation Center, Óbuda University, Budapest, Hungary.}
}



\maketitle
\thispagestyle{firstpage}
\begin{abstract}
Surgical teleoperation deprives surgeons of touch, forcing visual inference of tissue forces; the commercial platforms now restoring force feedback remain closed, preventing open research on haptic shared control. This \rev{paper provides a system-contribution: it presents an open-source, ROS\,2-native framework that integrates the compact, three-degrees-of-freedom (3-DOF) Haply Inverse3 haptic device with the da~Vinci Research Kit (dVRK). The framework includes a novel mapping strategy that combines 3-DOF translational device input with separately tracked VerseGrip orientation into 6-DOF patient-side manipulator commands.} Two virtual-fixture-inspired guidance modes serve as \rev{reference implementations to evaluate the framework} in a within-subjects user study on a teleoperated hot-wire benchmark task in simulation, and the architecture is further demonstrated on physical dVRK hardware. Relative to unassisted manual control, linear guidance significantly reduced both the number of wire contacts and the task completion time, whereas the progressive-stiffness \rev{guidance} produced consistent but statistically non-significant trends in the same direction. Subjective ratings of workload, agency, and satisfaction showed \rev{favorable descriptive trends} for the guidance modes. The results demonstrate the feasibility of haptic shared control with a compact device coupled to the dVRK and provide an openly available, reproducible platform for future research. They further suggest that simpler linear \rev{feedback} may be more robust for inexperienced operators than more complex profiles.
\end{abstract}


\section{Introduction}
\label{sec:intro}
Robotic surgical systems such as the da Vinci platform have transformed minimally invasive surgery, but a fundamental limitation remains: the physical separation between surgeon and patient removes natural tactile and force feedback. Current clinical systems have therefore relied primarily on visual cues and learned motor strategies to regulate interaction forces~\cite{Okamura2009,Kitagawa2005SensorySubstitution}.
This sensory gap can increase applied forces, elevate the risk of tissue damage, and make it harder to discriminate tissue properties during manipulation~\cite{Bergholz2023,Patel2022}.
Although da Vinci~5 introduces force feedback, the platform remains closed, which limits experimental access to the feedback architecture, custom control designs, and alternative haptic rendering strategies.
Recent studies suggest that force feedback can reduce applied tissue forces and improve task performance, but the available evidence is still limited and does not yet isolate force feedback from all other platform changes in clinical use~\cite{Awad2024,Servais2025,Chang2025,Kneuertz2025}.
Moreover, because human-in-the-loop control is a complex, multidimensional phenomenon, the design of effective haptic shared control strategies for surgical teleoperation remains an open question that requires further investigation~\cite{Grobbel_SMC2023,VFlemisch_2024_AT,Varga2024LQSharedControl}.
In contrast, research platforms such as the da Vinci Research Kit (dVRK) provide the openness needed to investigate new control architectures and human-robot interaction concepts~\cite{Kazanzides2014,DEttorre2021}.

\rev{Integrating haptic devices, such as the 3-DOF Haply Inverse3, with the 6-DOF dVRK requires combining secondary orientation sensing with translation and matching coordinate frames.}
\rev{The primary goal of this work is the system integration itself: providing the research community with an open, ROS\,2-native solution for using compact haptic devices with the dVRK. The novelty lies in the end-to-end integration of a compact 3-DOF haptic device with a da~Vinci-derived teleoperation platform. The haptic shared-control evaluation presented in this paper serves as a means of validating this platform rather than as an end in itself; the evaluated feedback modes are intentionally simple implementations.} The hot-wire task is used as a benchmark because it emphasizes path following, speed regulation, and collision avoidance, all of which are relevant to surgical teleoperation. 

In summary, the contributions of this paper are as follows:
\begin{enumerate}
    \item We develop an open-source ROS\,2 architecture \rev{that enables teleoperation between the Haply Inverse3 with the VerseGrip and the dVRK, including a specified mapping pipeline.} \newpage
    \item \rev{We implement two virtual-fixture-inspired \rev{guidance modes (linear and progressive-stiffness)} as reference implementations within this framework, rendered under matched maximum forces.}
    \item We empirically evaluate human-in-the-loop performance in a controlled user study, \rev{showing improvement in both wire contacts and completion time under linear guidance, and demonstrate the transferability of the architecture to physical dVRK hardware.}
\end{enumerate}

\section{State of the Art}
\label{sec:sota}

This section reviews the state of the art across the five relevant interrelated themes and identifies the research gaps that motivate the present work.

\subsection{Haptic Feedback in Robot-Assisted Surgery}
\label{sec:sota:haptics}
 
Early controlled experiments demonstrated that restoring force feedback to the operator in robot-assisted surgery can reduce tissue-damage errors by a factor of three~\cite{Okamura2004}, establishing the fundamental value proposition of haptics in surgical robotics. On the other hand, some studies suggest that 3D stereo vision in da Vinci systems can effectively replace haptic feedback with sufficient training~\cite{hagen2008visual}.
Subsequent work examined force sensing, display technologies, and the persistent engineering tradeoffs that continue to shape the field~\cite{Okamura2009}, while systematic analyses identified the stability--transparency tradeoff as the central bottleneck~\cite{Enayati2016}.
The instrumentation challenges -- miniaturization, biocompatibility, sterilizability, and cost -- that explain why earlier commercial systems lacked haptic feedback have been comprehensively surveyed~\cite{Patel2022}.
A meta-analysis synthesizing 56~studies confirmed large effect sizes for haptic feedback across multiple performance dimensions~\cite{Bergholz2023}.
Despite this strong evidence base, the question of how to move beyond passive force rendering toward cooperative guidance remains open.
 
\subsection{From Haptic Feedback to Haptic Shared Control}
\label{sec:sota:shared}
 
A conceptual shift occurred when the research community moved beyond passively reflecting environmental forces and toward actively guiding the operator through cooperative control.
Haptic shared control was formalized as a paradigm in which guiding forces at the master device encode a desired trajectory while the human retains override authority~\cite{Abbink2012}; experimental evidence showed that such guidance yields faster, more accurate control with lower effort and reduced visual attention demand.
A comprehensive taxonomy later distinguished traded, cooperative, shared, and supervisory control paradigms and provided general axioms for design and evaluation across several domains, including surgery~\cite{Abbink2018}.
 
The benefits of haptic guidance during telemanipulation were experimentally validated even when direct environment feedback is limited~\cite{Boessenkool2013}, positioning shared control as a viable alternative to improving haptic transparency alone.
However, guidance inaccuracies can degrade performance when model errors are significant~\cite{VanOosterhout2015} - a critical consideration for surgery where anatomical variability limits guidance-model fidelity.
 
In the surgical domain, a dual-user framework for haptic telesurgical training with a variable authority factor and $H_\infty$ robust stability guarantees was proposed early on~\cite{Nudehi2005}. Haptic shared control was subsequently extended to dual-arm telerobotic manipulation with passivity-based stability analysis~\cite{Selvaggio2018HSC}, and a broader survey situated these efforts within the wider autonomy landscape of physical human--robot interaction~\cite{Selvaggio2021}.
 
The dominant mechanism for enforcing shared control in surgical robotics is the virtual fixture, originally introduced as a computer-generated perceptual overlay that constrains or guides telerobotic manipulation~\cite{Rosenberg1993,2025_SharedControlApproach_varga}.
Guidance virtual fixtures decompose forces into preferred and non-preferred directions, while forbidden-region virtual fixtures prevent workspace boundary violations; the analytical framework addressing stability and passivity for both types remains foundational~\cite{Abbott2007}.
A comprehensive survey consolidated the field by covering constraint definition, evaluation, and enforcement across domains~\cite{Bowyer2014}.
Ensuring stable interaction under time-varying impedance has been addressed through energy-tank-based control architectures~\cite{Ferraguti2015}, extended to passive task-prioritized shared control with haptic guidance~\cite{Selvaggio2019ICRA}.
On the dVRK, adaptive virtual fixtures with online parameter generation have been demonstrated~\cite{Selvaggio2018VF}. These approaches provide a mature control-theoretic toolkit, but their application has been tied to specific device--platform combinations.

\subsection{Haptic Interfaces for Surgical Teleoperation}
\label{sec:sota:devices}
 
The quality of haptic shared control is fundamentally constrained by the master device through which the surgeon perceives and generates forces.
The Force~Dimension sigma.7 -- a 7-DOF device with 20\,N Cartesian force capability, gravity compensation, and fully actuated translation, rotation, and grasping -- set the standard~\cite{Tobergte2011} and has become the most widely used research-grade haptic master for surgical teleoperation.
A direct comparison of the sigma.7, the 3D~Systems HD\textsuperscript{2}, and the SensAble PHANToM~Premium~3.0 showed that device kinematics resembling the human upper extremity improve teleoperation performance~\cite{Zareinia2015}.
Kinematic redundancy in haptic masters reduces apparent inertia and improves soft-tissue stiffness discrimination~\cite{Torabi2019}, while cutaneous and tactile interfaces offer stability advantages over grounded kinesthetic devices~\cite{Pacchierotti2024}.
Despite this landscape, the set of haptic devices actually integrated with surgical teleoperation platforms remains remarkably narrow, dominated by the sigma.7 and the PHANToM family.
More generally, using generic 6-DOF devices as human--machine interfaces can be challenging for task-specific robotic applications because their kinematics are not necessarily matched to those of the controlled robot~\cite{10161348}.
Compact 3-DOF haptic devices may therefore offer a more adaptable alternative for selected surgical teleoperation tasks, motivating the present work, which integrates and evaluates such a device with a surgical robotic platform.

\subsection{User Study Design and Evaluation Methodology}
\label{sec:sota:eval}
 
Rigorous evaluation of haptic shared control in surgical contexts demands standardized tasks, validated instruments, and careful experimental design -- yet the literature reveals considerable inconsistency in tasks, metrics, and participant populations~\cite{Bergholz2023}.
A single-blinded randomized controlled study with residents performing both simulated and live bariatric surgeries demonstrated significantly lower NASA Task Load Index (NASA-TLX) workload scores with haptic feedback~\cite{Gomez2025}.
Sensorless haptic feedback on the dVRK has been evaluated in palpation and peg-transfer tasks using NASA-TLX, interaction forces, and completion times~\cite{Yilmaz2024Enhancing}, while the combination of haptic feedback with haptic guidance has been assessed using both NASA-TLX and the System Usability Scale~\cite{Schleer2020}.
Multi-modal feedback on the da~Vinci IS1200 achieved approximately 50\% grip force reduction~\cite{Abiri2019}, and skin-deformation tactile feedback was compared against force feedback and combined modalities~\cite{Quek2019}.
 
While these studies represent methodological best practice, evaluation tasks vary widely, making cross-study comparison difficult.
The hot-wire task -- a well-established precision benchmark in haptics and teleoperation with a clear binary error metric and straightforward completion-time measurement -- has not yet been employed as a benchmark for haptic shared control in surgical teleoperation.

\subsection{The da~Vinci Research Kit as Integration Platform}
\label{sec:sota:dvrk}
 
The dVRK repurposes retired first-generation da~Vinci hardware with custom field-programmable gate array-based electronics and open-source software, providing researchers at over 40~institutions with complete access to all control levels of a clinical-grade surgical robot~\cite{Kazanzides2014}.
A comprehensive review catalogued the first decade of dVRK research spanning haptic feedback, bilateral teleoperation, virtual fixtures, force sensing, and external haptic device integration~\cite{DEttorre2021}.
 
Closing the haptic loop requires either adding sensors or estimating interaction forces. A non-invasive trocar-embedded sensor was proposed that measures instrument-tissue forces without modifying the surgical instrument~\cite{Fontanelli2020}, while sensorless four-channel bilateral teleoperation using deep neural networks for dynamics identification has been shown to achieve high-performance haptic teleoperation without hardware modifications~\cite{Yilmaz2024Sensorless}.
 
On the master side, the dVRK's original master tool manipulators have been replaced with sigma.7 devices, and a user study demonstrated that force reflection significantly improves palpation and incision performance~\cite{Saracino2019}. Published studies coupling an external haptic device to the dVRK remain very limited. Two observations follow: the dVRK ecosystem remains tightly coupled to a small set of established haptic devices without a ROS\,2-native modular integration, and the vast majority of dVRK haptic studies focus on rendering environment forces rather than providing cooperative guidance forces in a shared-control paradigm.

\subsection{Research Gaps}
\label{sec:gaps}

The literature shows substantial progress in haptic feedback and shared-control methods, yet critical gaps remain. First, dVRK studies are dominated by established master devices like the sigma.7 and PHANToM family~\cite{Saracino2019,Yilmaz2024Sensorless,Zareinia2015}, while compact devices such as the Haply Inverse3 lack systematic integration. Moreover, existing integrations rely on legacy middleware rather than modular ROS\,2-native architectures. Second, most research focuses on reflecting environment forces for transparency rather than cooperative haptic guidance~\cite{Yilmaz2024Enhancing,Saracino2019,Schleer2020}. Although mature control-theoretic tools exist~\cite{Abbink2012,Bowyer2014,Abbott2007,Selvaggio2018VF}, their application to real-time assistance with compact haptic devices remains insufficiently explored. Third, evaluation methodology is fragmented across diverse tasks, complicating cross-study comparison~\cite{Bergholz2023}. Specifically, the hot-wire task has not been systematically used to evaluate haptic shared control. Collectively, these gaps indicate a need for a unified framework combining accessible hardware, cooperative strategies, and standardized benchmarks. 

These limitations of the state of the art motivate and necessitate this work. The following sections detail the system architecture and experimental results required to bridge these identified limitations.

\section{Overall System Architecture}
\label{sec:system}
\begin{figure*}
    \centering
    \includegraphics[width=0.9\textwidth]{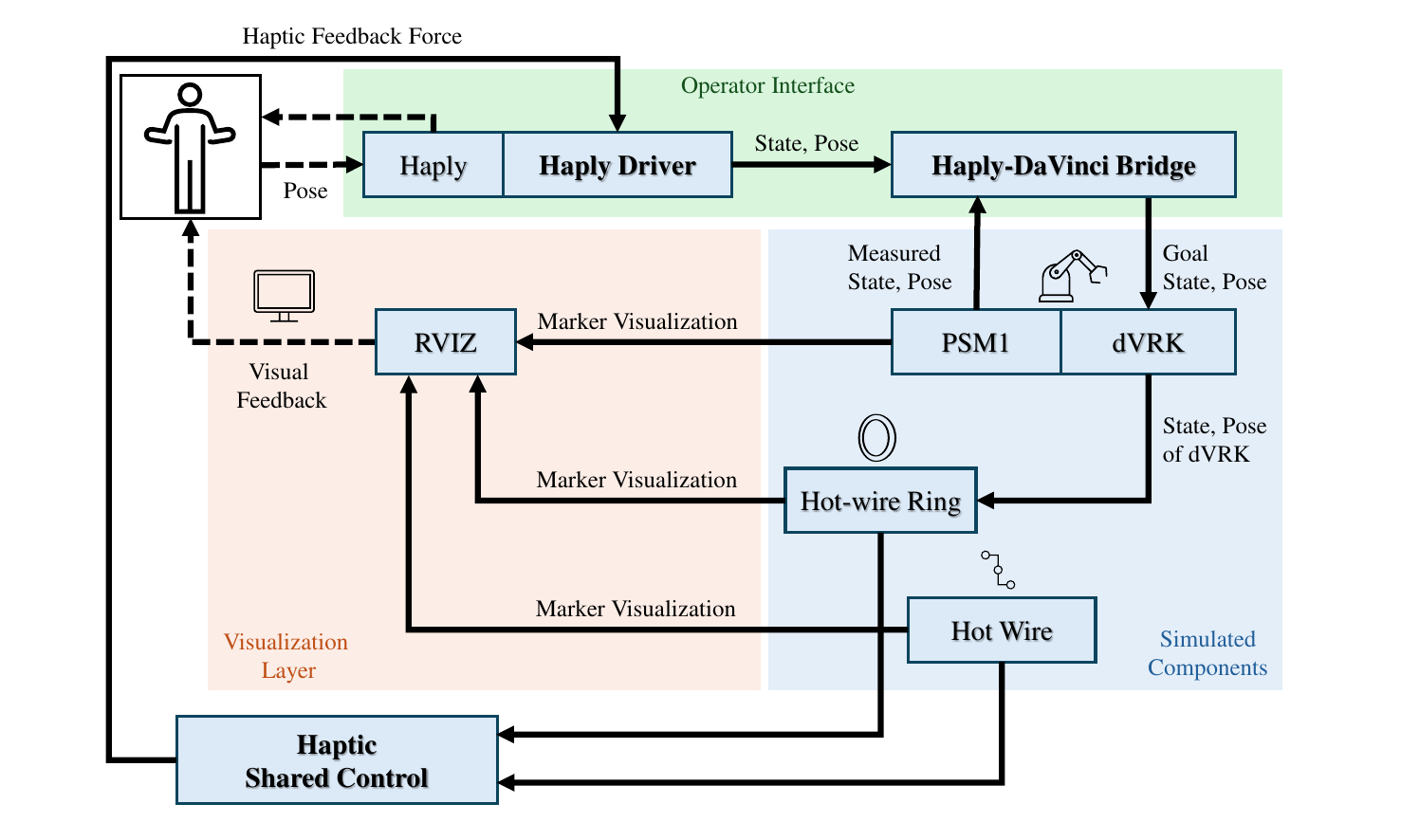}
    \caption{Overall system architecture of the proposed haptic shared-control framework. The Haply Driver reads the state of the Haply interface and transmits it to the Haply–DaVinci Bridge, where mapping and shared-control logic are applied. The resulting goal pose is sent to the dVRK/PSM1, and the measured robot state is fed back to the bridge and the visualization layer. RViz renders the marker-based visualization of the hot-wire task environment. Bold blocks represent the software modules developed in this work.}
    \label{fig:architecture}
\end{figure*}

We developed a software architecture that integrates the dVRK with the Haply Inverse3 haptic device. In addition, a mapping strategy was implemented to enable haptic shared control of the dVRK using the Inverse3 equipped with the VerseGrip orientation tracking addon, allowing for both force-based and position-based control modes. The overall system architecture is illustrated in Figure~\ref{fig:architecture}.

\subsection{Haply Inverse3 and VerseGrip ROS\,2 Package}
Since the manufacturer does not provide a native ROS\,2 interface, and only exposes hardware communication through a WebSocket-based service API, we developed a package that wraps this interface and enables real-time reading of device position and orientation, position- and force-based control, and visualization through standard ROS\,2 topics.

The driver node establishes a persistent interface with the \texttt{haply-inverse-service} utilizing the WebSocket protocol, facilitating the continuous acquisition of real-time telemetry. The temporal resolution of incoming spatial data depends on the native sampling frequencies of the hardware peripherals. Upon ingestion, the kinematic state data-comprising the Cartesian position and linear velocity of the Inverse3, alongside the quaternion orientation and discrete input states of the VerseGrip Stylus.

The system architecture supports dual control modalities through the \texttt{haply\_target} topic:
\begin{itemize}
    \item \textbf{Force-based Control \rev{Modality}:} Commanded force vectors are applied directly to the Inverse3 end-effector.
    \item \textbf{Position-based Control \rev{Modality}:} The driver derives corrective forces via a software-implemented Proportional-Integral-Derivative (PID) controller.
\end{itemize}
Command transmission is synchronized with the reception of WebSocket messages; thus, the effective control loop frequency is functionally dependent on the update rate of the \texttt{haply-inverse-service}.

\rev{Operational safety is maintained through complementary mechanisms with different response times. (1) During initialization, the driver node executes a diagnostic verification of all connected hardware, logging device identifiers, calibration status, and functional readiness. (2) During operation, output forces are subject to a configurable per-axis saturation limit, and a command watchdog sets force outputs to zero if no control command is received within $\tau_{\mathrm{wd}}= 0.5$ s. These mechanisms were active in all experimental runs reported in this paper.}

\rev{A separate connection-loss mechanism performs an orderly node shutdown if hardware communication is unavailable for longer than the configurable timeout $\tau_{\mathrm{disc}} = 100 \cdot \tau_{\mathrm{wd}} = 50$ s. As the command watchdog already sets force output to zero upon stale commands, this timeout serves process liveness rather than immediate force safety.}

To enable real-time visual diagnostics during experimental procedures, a dedicated \texttt{rviz\_visualization\_node} monitors the device state and publishes mesh-based 3D markers to the RViz environment, providing an accurate representation of the device's pose and orientation.
The package further includes a lightweight mesh model of the VerseGrip Stylus for visualization, along with custom ROS\,2 message definitions that support device state representation and control communication. Additionally, a set of demonstration nodes is provided for validation and evaluation. These comprise a sinusoidal motion demo, illustrating position-based control through periodic single-axis trajectories, and a haptic ball demo, which emulates a virtual spherical constraint generating restoring forces upon boundary penetration, thereby demonstrating force-based interaction. Subscriber nodes are also available to enable detailed logging of device state data.

The package is released as open-source software under the MIT License and is publicly available on GitHub\footnote{\url{https://github.com/ABC-iRobotics/haply_ros2_interface}} facilitating the integration of the Haply Inverse3 and VerseGrip Stylus into other ROS 2-based platforms and experiments.

\subsection{The Proposed Mapping Approach for VerseGrip and dVRK Integration}
\label{sec:mapping}
\rev{The Inverse3 tracks 3-DOF stylus position and provides force feedback, while the VerseGrip supplies orientation and buttons. The Haply--dVRK bridge combines these into 6-DOF commands for the PSM1. Two mapping variants are implemented: Cartesian (commanding end-effector pose) and joint-space (commanding PSM joints directly).}

\subsubsection{Reference States and Increments}
\rev{The proposed mapping in our framework is relative. Once a valid robot state is available, the bridge stores the Haply position ${}^{H}\vek{p}_{h,0}$, the stylus roll-pitch-yaw angles $\sv{\eta}_{0}=[\phi_{0},\theta_{0},\psi_{0}]^{\mathrm{T}}$ extracted from the VerseGrip quaternion, and the measured robot state as references. All subsequent commands are computed from the increments}
\begin{align}
    \rev{\Delta\vek{p}} &\rev{= s_p\,\vek{C}_p\left({}^{H}\vek{p}_h-{}^{H}\vek{p}_{h,0}\right),} \\
    \rev{\vek{C}_p} & =\operatorname{diag}(-1,\,1,\,-1), \label{eq:dp}\\
    \rev{\Delta\sv{\eta}} &\rev{= [\Delta\phi,\Delta\theta,\Delta\psi]^{\mathrm{T}} = s_r\,\operatorname{wrap}\!\left(\sv{\eta}-\sv{\eta}_{0}\right),} \label{eq:deta}
\end{align}
\rev{where $s_p=0.25$ is the translational scaling factor and $s_r$ the rotational scaling factor. The matrix $\vek{C}_p$, a rotation by $\pi$ about the common $y$-axis, aligns the Haply axes with the PSM base frame. The operator $\operatorname{wrap}(\cdot)$ maps the roll and pitch differences to $(-\pi,\pi]$, while the yaw angle is unwrapped to permit continuous stylus rotation.}

\subsubsection{Cartesian Mapping}
\rev{As an alternative configuration, the bridge also supports a Cartesian mapping. With the reference pose $({}^{R}\vek{p}_{E,0},\,\phi_{E,0},\,\theta_{E,0},\,\psi_{E,0})$ obtained from \texttt{measured\_cp}, the goal pose is}
\begin{align}
    \rev{{}^{R}\vek{p}_{E,d}} &\rev{= {}^{R}\vek{p}_{E,0} + \Delta\vek{p},} \label{eq:cart_pos}\\
    \rev{{}^{R}\vek{R}_{E,d}} &\rev{= \vek{R}_{\mathrm{RPY}}\!\left(\phi_{E,0}-\Delta\theta,\; \theta_{E,0}-\Delta\phi,\; \psi_{E,0}-\Delta\psi\right),} \label{eq:cart_rot}
\end{align}
\rev{where $s_r=0.5$. The exchange of roll and pitch and the sign inversions align the stylus axes with the instrument axes. In this variant, the pose is sent through the non-interpolated Cartesian interface. This variant is provided for completeness but was not used in the experiments reported in this paper.}

\subsubsection{Joint-Space Mapping}
\rev{With the measured reference configuration $\vek{q}_0$ and the PSM joint order $\vek{q}=[q_1,\dots,q_6]^{\mathrm{T}}$ (outer yaw, outer pitch, insertion, instrument roll, wrist pitch, wrist yaw), the goal configuration is}
\begin{equation}
    \rev{\vek{q}_d = \vek{q}_0 + \left[\,g\,\Delta p_x,\; g\,\Delta p_y,\; \Delta p_z,\; -\Delta\psi,\; \Delta\phi,\; \Delta\theta\,\right]^{\mathrm{T}},}
    \label{eq:joint_map}
\end{equation}
\rev{with $s_r=1$ and $g=\SI{10}{\radian\per\meter}$. Thus, the instrument roll follows the stylus yaw, and the wrist pitch and wrist yaw follow the stylus roll and pitch, respectively. Since the lateral tip displacement about the remote center of motion scales approximately with $q_3\,\Delta q_{1,2}$, the gain $g$ yields lateral and axial motion scaling of similar magnitude at insertion depths of about \SI{0.1}{\meter}. The goal configuration is sent through the interpolated joint interface to obtain smooth motion.}

\rev{In all experiments reported in this work, the joint-space mapping of \eqref{eq:joint_map} was used, commanding the PSM through the interpolated interface.}

\subsubsection{Clutching, Gripper Control, and Force Mapping}
\rev{Pressing VerseGrip button~B resets the translational references, i.e., ${}^{H}\vek{p}_{h,0}$ together with ${}^{R}\vek{p}_{E,0}$ or $q_{1,0},q_{2,0},q_{3,0}$, while the orientation references are kept. This allows the operator to recenter the Inverse3 within its workspace without moving the instrument. A rising edge of button~A toggles the gripper between open (\SI{1}{\radian}) and closed (\SI{0}{\radian}).}

\rev{The feedback forces of Section~\ref{sec:shared_control} are computed in the task frame $\{W\}$ of the hot-wire environment. Since the Haply cursor frame $\{H\}$ and $\{W\}$ differ only by a fixed axis convention, positions, velocities, and forces are converted between the two frames through the constant transform $\vek{S}_z=\operatorname{diag}(1,\,1,\,-1)$:}
\begin{equation}
    \rev{{}^{W}\vek{p} = \vek{S}_z\,{}^{H}\vek{p}_h, \quad
    {}^{W}\vek{v} = \vek{S}_z\,{}^{H}\vek{v}_h, \quad
    {}^{H}\vek{f}_h = \vek{S}_z\,{}^{W}\vek{f}_g.}
    \label{eq:force_map}
\end{equation}
\rev{Because $\vek{S}_z$ is orthogonal and self-inverse, \eqref{eq:force_map} preserves both the force magnitude specified by the guidance and the exchanged mechanical power, ${}^{H}\vek{f}_h^{\mathrm{T}}\,{}^{H}\vek{v}_h = {}^{W}\vek{f}_g^{\mathrm{T}}\,{}^{W}\vek{v}$. The guidance distance and normal are likewise computed from the converted position ${}^{W}\vek{p}$.}

\section{Design of the Reference Guidance Modes}
\label{sec:shared_control}
In the following, we present two guidance modes for the hot-wire task: M1, which provides linear guidance based on the distance to the desired trajectory, and M2, which increases the stiffness of the feedback force as the user approaches the target path. 
\subsection{Linear and Progressive Stiffness Characteristics}
\subsubsection{Mode~1} The linear force feedback mode (M1) applies a restoring force proportional to the effective penetration distance $d$, measured from the inner boundary of the haptic zone $d_i$ to the wire surface. The maximum force is limited by the outer effective radius $d_o$. The force increases monotonically as the operator moves further into the zone, providing continuous guidance back toward the wire center:
\begin{equation} \label{eq:linear_feedback_M1}
    \mathbf{F} = k\, d\,\hat{\mathbf{n}} - b\, \mathbf{v}.
\end{equation}
where $\mathbf{F}$ is the force applied to the user, $k$ is a constant stiffness coefficient, and $d$ is the distance from the desired trajectory. The vector $\hat{\mathbf{n}}$ is the unit normal pointing from the current end-effector position toward the closest point on the desired trajectory, ensuring that the feedback force is always directed to guide the operator back toward the target path.
The damping term $-b\,\mathbf{v}$ acts on the measured end-effector velocity $\mathbf{v}$ and serves to suppress oscillations that would otherwise arise from the elastic restoring force alone.\\
\begin{figure}[!t]
    \centering
    \includegraphics[width=0.494\textwidth]{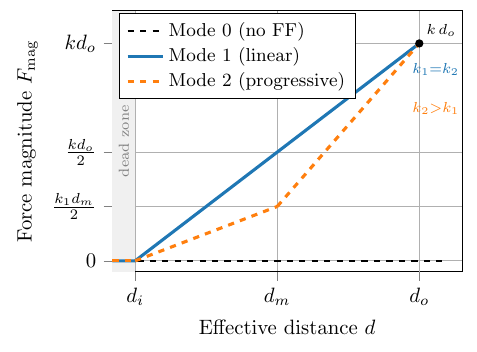}
    \caption{Force feedback characteristics for the two haptic modes.}
    \label{fig:ff_characteristics}
\end{figure}
\subsubsection{Mode~2} implements a piecewise-linear stiffness profile, motivated by the limitations of a uniform force field in precision teleoperation. A constant stiffness throughout the haptic zone introduces a trade-off between sensitivity near the wire surface and sufficient corrective authority at the zone boundary: a high uniform stiffness impedes fine manipulation, while a low one fails to provide adequate guidance as the operator approaches the boundary.

Mode~2 resolves this conflict by partitioning the haptic zone at a midpoint $d_m$ into two regions of distinct stiffness, rendering a compliant response in the inner region and an escalating corrective force in the outer region. The resulting force vector is given by 
\begin{equation} \label{eq:feedback_Mode2}
    \mathbf{F} = F_{\mathrm{mag}}(d)\,\hat{\mathbf{n}} - b\,\mathbf{v}, 
\end{equation} 
where the scalar magnitude is defined as 
\begin{equation}  \label{eq:feedback_ModesF}
    \! \! F_{\mathrm{mag}}(d) = 
    \begin{cases} k_1\,(d - d_i) & \text{if } d \leq d_m, \\[4pt] 
        k_1\,(d_m - d_i) + k_2\,(d - d_m) & \text{if } d > d_m, 
    \end{cases} 
\end{equation} 
with 
$k_1 = \tfrac{k}{2}$, $d_m = \tfrac{d_i + d_o}{2} \; \mathrm{and}$ 
\begin{equation} \label{eq:feedback_Modes_k2}
    k_2 = \frac{k\,d_o - k_1\,(d_m - d_i)}{d_o - d_m}. 
\end{equation} 
In the inner subzone ($d \leq d_m$), the reduced stiffness $k_1$ preserves operator dexterity and avoids premature force saturation during intentional near-wire maneuvers. As the operator penetrates beyond $d_m$, the stiffness transitions to $k_2 > k_1$, producing a perceptually salient increase in resistive force that communicates proximity to the zone boundary without discontinuity. 
The parameter $k_2$ is derived analytically such that $F_{\mathrm{mag}}(d_o)$ coincides with the force rendered by Mode~1 at the same location, ensuring identical maximum forces across both modes and thus enabling a controlled, fair comparison of the two guidance modes in the user study.

The implemented force feedback characteristics for both modes are illustrated in Figure~\ref{fig:ff_characteristics}, where the force magnitude is plotted against the distance from the desired trajectory.

\section{User Study}
\label{sec:user_study}

The user study is designed to evaluate the effectiveness of the proposed haptic shared control strategies under limited spatial visualization, since only a standard 2D monitor is used for visual
feedback. Two implementations of the shared control strategy are evaluated, referred to as M1 and M2. The manual control mode without active haptic guidance is denoted M0 and serves as the baseline.

\subsection{Working Hypotheses}
\label{sec:hypotheses}

The working hypotheses for the user study are as follows:
\begin{itemize}
    \item \textbf{H1:} The haptic shared control strategies (M1 and M2) lead to a significant reduction in the number of contacts with the wire compared to the manual control mode~(M0).
    \item \textbf{H2:} The haptic shared control strategies (M1 and M2) result in significantly shorter task completion times compared to the manual control mode~(M0). 
    \item \textbf{H3:} A haptic shared control strategy leads to significantly better subjective user experience ratings compared to the manual control mode~(M0): The cognitive workload, the sense of agency, and overall satisfaction are significantly improved under haptic shared control.
\end{itemize}

\subsection{Study Design}
\label{sec:design}
Twenty test subjects (7 female, 13 male) were recruited from a pool of volunteers, ranging in age from 25 to 65 years. Prior experience with teleoperation systems varied across participants, from none to advanced, providing a heterogeneous sample. The study followed a within-subjects design in which each participant completed the hot-wire task under all three control modes (M0, M1, M2). To mitigate learning effects, the order of modes was counterbalanced across participants using a balanced Latin square arrangement.

Prior to the actual runs, the participants underwent a five-minute training session for each mode, resulting in a total familiarization time of 15 minutes per participant before data collection began.

Visual feedback was provided throughout the task via a color change of the wire upon contact, indicating a collision event. No additional auditory or haptic warning was given at the moment of contact beyond the mode-specific haptic rendering already active in M1 and M2. Visual feedback was restricted to a standard 2D monitor, providing no stereoscopic depth cues to the participants.

It should be noted that the ring used in the hot-wire task may become disengaged from the wire during the trial. In such cases, no haptic feedback is rendered until the participant manually repositions the ring onto the wire, as the shared control forces depend on proximity to the wire geometry. This results in a temporary interruption of the haptic feedback loop and was handled consistently across all three control modes.

All participants took part voluntarily and provided written informed consent prior to participation. The study protocol was reviewed and approved by the ethics committee, approval no. 596. Participants were informed of the purpose and procedure of the study and of their right to withdraw at any time without consequence. No personally identifiable information was recorded, and all data were anonymized prior to analysis.

\rev{The simulated study used the same software stack as the physical system in Section~\ref{sec:discussion_study}: the Haply driver, the Haply--dVRK bridge with the joint-space mapping, the guidance modes of Section~\ref{sec:shared_control}, and the PSM1 command interface. Only the physical PSM and the endoscopic camera were replaced, by a kinematic model of the PSM and a virtual hot-wire environment rendered in RViz. The study therefore evaluates the complete integration chain except the PSM dynamics and the physical camera.}

\subsection{Experimental Setup Details}
\label{sec:setup_details}
The hot-wire benchmark used in the user study is defined by the following specifications.  
The Haply\,Inverse3 haptic device is capable of generating forces between $0$\,N and $1.2$\,N, see \cite{HaplyInverse3}; this theoretical range informed the selection of guidance stiffness values for the shared-control strategies.  

The wire itself has a total length of $\color{blue}0.21$\,m and is divided into five straight segments. The first, second, third, fourth, and fifth segments measure $0.04$\,m, $0.03$\,m, $0.05$\,m, $0.05$\,m, and $0.04$\,m, respectively. These dimensions that reproduce the physical prototype illustrated in Fig.~\ref{fig:dv_integration}. A guiding ring attached to the wire possesses a diameter of $0.02$\,m.\\  
Control-mode parameters are defined as follows. 
\begin{itemize}
  \item \rev{Mode\,0 (baseline): Manual control without guidance forces, i.e., $\mathbf{F} = \mathbf{0}$.}
  \item \rev{Mode\,1 (linear guidance): Restoring force according to~\eqref{eq:linear_feedback_M1} with constant stiffness $k = 80$\,N/m and viscous damping coefficient $b = 2$\,N\,s/m.}
  \item \rev{Mode\,2 (progressive-stiffness guidance): Restoring force according to~\eqref{eq:feedback_Mode2}-\eqref{eq:feedback_Modes_k2} and  with piecewise-linear stiffness, i.e., $k_1 = 40$\,N/m in the inner region and $k_2 = 120$\,N/m in the outer region, and the same viscous damping coefficient $b = 2$\,N\,s/m.}
\end{itemize}

The virtual-fixture zone is bounded by an inner radius $d_{i}=0$, an outer radius $d_{o}=0.015$\,m, and a midpoint $d_{m}=(d_{i}+d_{o})/2$. The stiffness values $k_{1} =40$\,N/m and $\color{blue} k_{2} =120$\,N/m are derived analytically to guarantee that the maximum force at the outer boundary matches the linear guidance force of Mode\,1.

These numeric values ensure comparable maximum forces across the two guidance strategies while allowing distinct force-field characteristics within the haptic zone.   

\subsection{Measures}
\label{sec:measures}

The objective measures recorded for each participant and control mode are:

\begin{itemize}
    \item \textbf{Number of Contacts:} The total number of contacts between the tool and the wire during the task execution.
    \item \textbf{Task Completion Time:} The total time taken to complete the task, measured in seconds.
\end{itemize}

The following three questions are included in the post-task questionnaire to assess subjective user experience and perceived performance:

\begin{itemize}
    \item Q1: I found the task to be demanding. (Cognitive workload)
    \item Q2: I felt in full control of what I was doing. (Sense of agency)
    \item Q3: I am satisfied with the result of my work. (Overall satisfaction)
\end{itemize}

\section{Results}
\label{sec:results}
H1 and H2 were assessed using the two objective measures: number of contacts and task completion time. For each dependent variable, two planned pairwise comparisons were conducted (M0~vs.~M1 and M0~vs.~M2) using paired $t$-tests, as Shapiro--Wilk tests on the pairwise differences did not indicate departures from normality in all cases. The significance threshold was Bonferroni-adjusted to $\alpha = .025$ to account for the two comparisons. Descriptive statistics are summarized in Table~\ref{tab:descriptives} and inferential results in Table~\ref{tab:pairwise}.

For the subjective measures, responses to the three questions were analyzed using non-parametric Wilcoxon signed-rank tests for the same pairwise comparisons (M0~vs.~M1 and M0~vs.~M2), with a Bonferroni-adjusted significance threshold of $\alpha = .025$.

\begin{figure*}[t]
    \centering
    \includegraphics[width=0.85\textwidth]{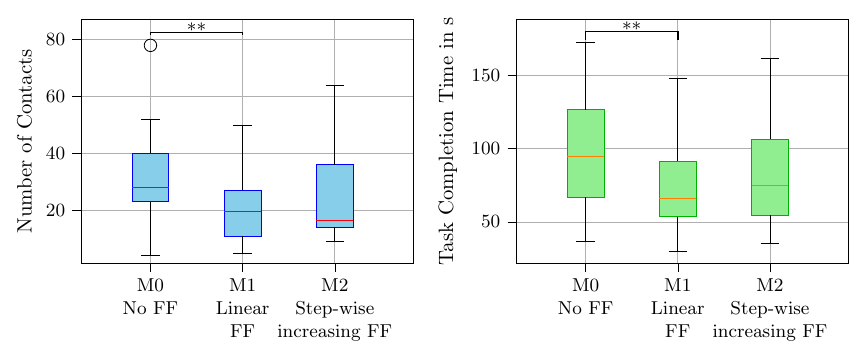}
    \caption{Left: Number of contacts with the wire per participant for the three control modes. Lower values indicate better performance. \\
    Right: Task completion time in seconds per participant for the three control modes. Lower values indicate better performance.}
    \label{fig:contacts_time}
\end{figure*}

\begin{table}[t]
\small
\centering
\caption{Descriptive statistics (mean and standard deviation) for the
         two objective measures across the three control modes
         ($N = 20$). \rev{The mean values with standard deviations in parentheses are reported.}}
\label{tab:descriptives}
\begin{threeparttable}
\begin{tabular}{lccc}
\thead{Measure}\! & \thead{M0} & \thead{M1} & \thead{M2} \\
\hline
Contacts\!        & \!31.85 (16.78)\!\! & \!20.90 (12.86)\!\! & \!24.35 (15.69)\!\! \\
Time (\si{\second})\! & \!96.31 (37.47)\!\! & \!75.01 (33.06)\!\! & \!84.25 (36.26)\!\! \\
\hline
\end{tabular}
\end{threeparttable}
\end{table}
\begin{table}[t]
\small
\centering
\caption{Pairwise paired $t$-test results for both objective measures.
         Bonferroni-corrected significance threshold: ${\alpha = 0.025}$.}
\label{tab:pairwise}
\begin{threeparttable}
\begin{tabular}{llcccl}
\thead{Measure} & \thead{Comparison} & \thead{$t(19)$} &
\thead{$p$} & \thead{$d$} & \thead{Sig.} \\
\hline
Contacts & M0 vs.\ M1 & 3.18 & .005 & 0.71 & $**$ \\
Contacts & M0 vs.\ M2 & 1.58 & .130 & 0.35 & ns \\
Time     & M0 vs.\ M1 & 3.45 & .003 & 0.77 & $**$ \\
Time     & M0 vs.\ M2 & 1.74 & .098 & 0.39 & ns \\
\hline
\end{tabular}
\begin{tablenotes}
\item $d$ = Cohen's $d$ (effect size). ns = not significant; $**$ $p < .025$ (Bonferroni-corrected). A positive $t$-value indicates M0 is worse (more contacts or longer time) than the comparison mode.
\end{tablenotes}
\end{threeparttable}
\end{table}

\subsection{Objective Measures: Number of Contacts}
Descriptive statistics revealed mean contact counts of $M = 31.85$ ($SD = 16.78$) for M0, $M = 20.90$ ($SD = 12.86$) for M1, and $M = 24.35$ ($SD = 15.69$) for M2 (see also left side of Figure~\ref{fig:contacts_time}). Participants produced significantly fewer contacts under M1 than M0, $t(19) = 3.18$, $p = .005$, $d = 0.71$ (medium-to-large effect). The difference between M0 and M2 did not reach significance, $t(19) = 1.58$, $p = .130$, $d = 0.35$, although the trend was in the same direction, with a mean reduction of $7.50$ contacts.

\subsection{Objective Measures: Completion Time}
Mean completion times were $M = 96.31$\,\si{\second} ($SD = 37.47$\,\si{\second}) for M0, $M = 75.01$\,\si{\second} ($SD = 33.06$\,\si{\second}) for M1, and $M = 84.25$\,\si{\second} ($SD = 36.26$\,\si{\second}) for M2 (see right side of Figure~\ref{fig:contacts_time}). Participants completed the task significantly faster under M1 than M0, $t(19) = 3.45$, $p = .003$, $d = 0.77$ (medium-to-large effect), corresponding to a mean reduction of $21.30$\,\si{\second}. The comparison between M0 and M2 did not reach the corrected threshold, $t(19) = 1.74$, $p = .098$, $d = 0.39$. Although the descriptive trend favored shorter completion times under M2, this difference did not reach statistical significance.

\subsection{Subjective Measures}
The subjective responses were collected on a symmetric seven-point Likert scale ranging from $-3$ to $+3$. Since the data are ordinal, Wilcoxon Signed-Rank tests were used for all pairwise comparisons. Shapiro--Wilk tests were additionally performed on the pairwise differences; violations were detected for both Q2 comparisons (M0$-$M1 and M0$-$M2), further confirming the appropriateness of the non-parametric approach. Two planned comparisons were conducted per question (M0 vs.\ M1 and M0 vs.\ M2), with the significance threshold Bonferroni-adjusted to $\alpha = .025$. Descriptive statistics and test results are summarized in Tables~\ref{tab:likert_descriptives} and~\ref{tab:likert_tests}.

\subsubsection{Question 1}
Participants rated M0 consistently higher than both M1 and M2, with a median of $+2.0$ for M0 compared to $+1.0$ for both M1 and M2. \rev{The comparison M0 vs.\ M1 reached nominal significance before correction ($W = 15.5$, $p = .034$) with a large effect size ($r = .659$), but did not survive Bonferroni correction.} The comparison M0 vs.\ M2 did not reach significance ($W = 28.0$, $p = .065$, \rev{$r = .533$), although the effect size was large.}

\subsubsection{Question 2}
M0 received the lowest ratings (Mdn $= -1.0$), while M1 was rated highest (Mdn $= +1.0$) and M2 fell in between (Mdn $= +0.5$). Neither comparison reached significance (M0 vs.\ M1: $W = 33.0$, $p = .112$, \rev{$r = .450$, a medium effect}; M0 vs.\ M2: $W = 13.5$, $p = .070$, \rev{$r = .591$, a large effect}).

\subsubsection{Question 3}
M0 was rated lower than both M1 and M2 (Mdn $= -1.0$ vs.\ $0.0$ for both). \rev{The comparison M0 vs.\ M1 reached nominal significance before correction {($W = 38.5$, $p = .038$)} with a large effect size ($r = .550$), but did not survive Bonferroni correction.} The comparison M0 vs.\ M2 did not reach significance ($W = 22.5$, $p = .056$, $r = .571$), although the effect size was large.

\subsubsection{Summary of the Subjective Results}
\rev{No pairwise comparisons attained statistical significance after Bonferroni correction. However, a systematic directional trend was observed across the three questions: participants consistently rated the haptic conditions more favorably than the non-haptic baseline M0. The two comparisons M0 vs.\ M1 for perceived demand (Q1) and satisfaction (Q3) reached nominal significance ($p = .034$ and $p = .038$) prior to correction, and the effect sizes ranged from $r = .450$ to $r = .659$, corresponding to medium-to-large effects. Given the limited sample size ($N = 20$), these findings may reflect insufficient statistical power to detect small-to-moderate subjective effects after correction for multiple comparisons, and should therefore be interpreted as descriptive trends rather than confirmatory evidence.}

\begin{figure*}
    \centering
    \includegraphics[width=0.85\textwidth]{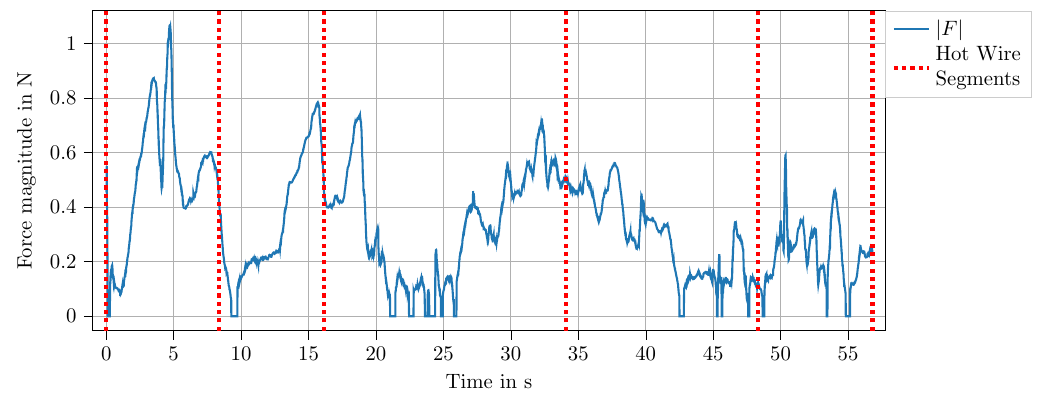}
    \caption{Example output forces generated by the Haply Inverse3 haptic device in case M1 was selected.}
    \label{fig:haply_output_forces}
\end{figure*}

To support the subjective findings with objective evidence, Fig.~\ref{fig:haply_output_forces} illustrates the force feedback produced by the Haply Inverse3 in a representative experiment using the Mode 1, linear feedback force method. The red lines indicate the endpoints of the hot-wire segments during the study, where the orientation of the hot-wire ring must have been adjusted in order to follow it. These results indicate that the control framework rendered guidance forces as designed; the subjective ratings therefore reflect operator perception rather than system-performance limitations.

\begin{table}[t]
\caption{Descriptive statistics for the subjective Likert responses ($-3$ to $+3$) across the three control modes ($N = 20$). \rev{The mean values with standard deviations in parentheses are reported.}}
\label{tab:likert_descriptives}
\centering
\begin{tabular}{cccc}
\toprule
 & \multicolumn{3}{c}{Control Mode} \\
\cmidrule(lr){2-4}
Question & M0 & M1 & M2 \\
\midrule
Q1 & $+1.85$ (0.93) & $+0.95$ (1.36) & $+1.05$ (1.39) \\
Q2 & $-0.30$ (1.59) & $+0.25$ (1.74) & $+0.15$ (1.53) \\
Q3 & $-0.95$ (1.39) & $+0.05$ (1.82) & $-0.15$ (1.69) \\
\bottomrule
\end{tabular}
\end{table}
\begin{table}[t]
\small
\caption{Wilcoxon Signed-Rank test results for pairwise comparisons
         of M0 against M1 and M2 per question ($N = 20$).
         Bonferroni-corrected significance threshold: $\alpha = .025$.}
\label{tab:likert_tests}
\centering
\begin{tabular}{llcccc}
\toprule
Comparison & $W$ & $p$ & $r$ & Sig. \\
\midrule
Q1: M0 vs.\ M1 & 15.5 & .034 & .659 & $*$ \\
Q1: M0 vs.\ M2 & 28.0 & .065 & .533 & ns \\
\hline
Q2: M0 vs.\ M1 & 33.0 & .112 & .450 & ns \\
Q2: M0 vs.\ M2 & 13.5 & .070 & .591 & ns \\
\hline
Q3: M0 vs.\ M1 & 38.5 & .038 & .550 & $*$ \\
Q3: M0 vs.\ M2 & 22.5 & .056 & .571 & ns \\
\bottomrule
\end{tabular}
\begin{tablenotes}
\small
\item $r$ = rank-biserial correlation (effect size);
      $|r| \geq .50$ large, $.30$--$.49$ medium, $< .30$ small.
      ns = not significant; $*$ $p < .05$ (uncorrected);
      following Bonferroni adjustment, all comparisons were non-significant. ($\alpha = .025$).
\end{tablenotes}
\end{table}

\begin{figure*}[t]
	\centering
	\includegraphics[width=0.7\textwidth]{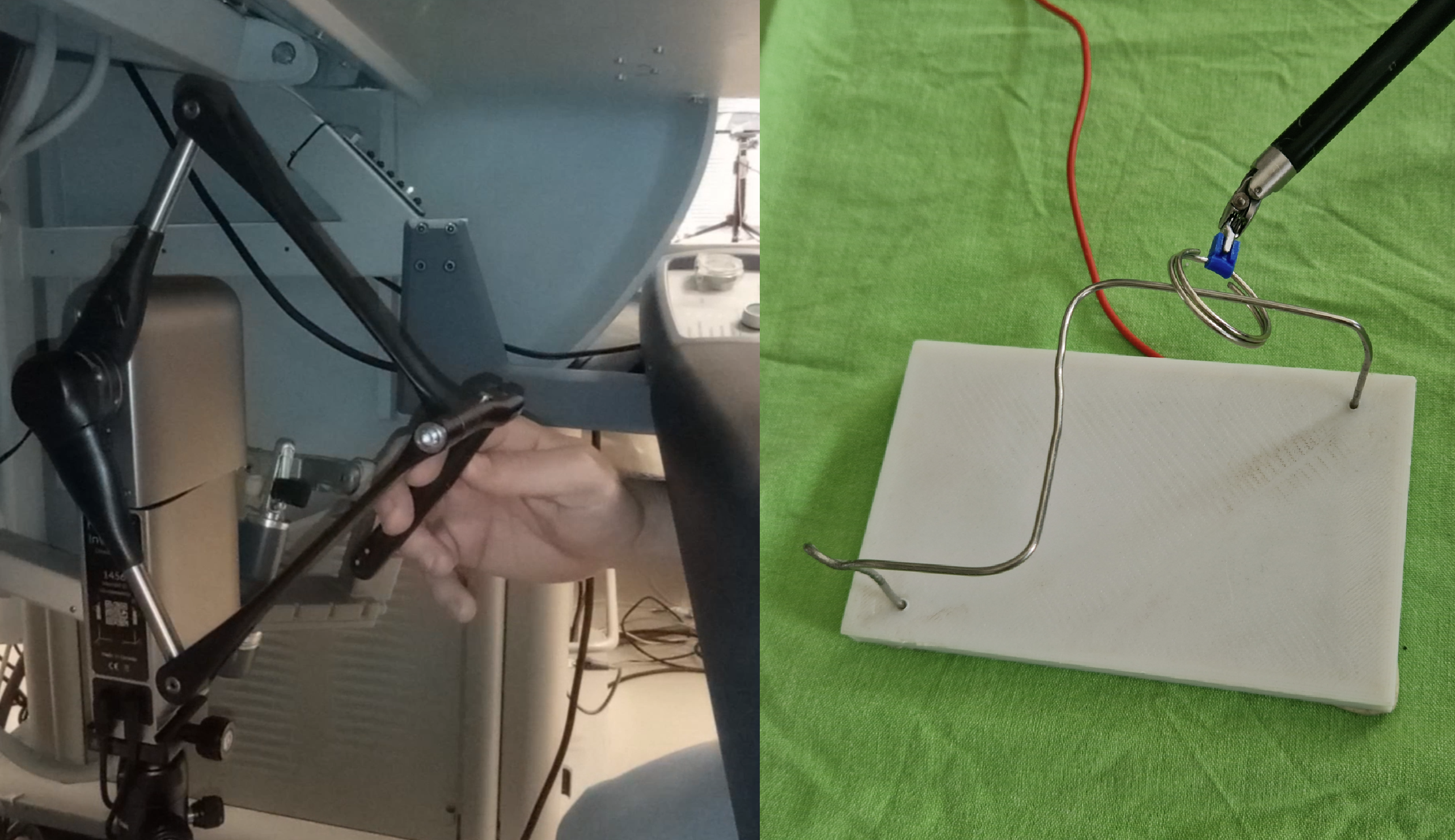}
	\caption{Integration of the dVRK with the Haply Inverse3 haptic device.}
	\label{fig:dv_integration}
\end{figure*}

\section{Discussion of the User Study Results}
\label{sec:discussion_study}

\subsection{Implications}
\label{sec:implications}
While Mode~1 demonstrated statistically significant improvements over the baseline in both objective measures, Mode~2 failed to reach the corrected significance threshold despite showing similar directional trends. 
\rev{This discrepancy suggests that the piecewise stiffness profile in Mode~2, while theoretically advantageous for balancing sensitivity and authority, may have introduced additional cognitive load. The lack of significance for Mode~2 cannot be attributed solely to statistical power; rather, it indicates that increased control complexity may negate the benefits of progressive guidance in novice operators. This aligns with the subjective reports, and suggests that simpler linear guidance may be more robust for users with limited teleoperation experience.}

\rev{Beyond the specific guidance profiles, these results primarily validate the platform: the integrated system rendered stable haptic guidance through a compact 3-DOF device coupled to the dVRK, and its open-source architecture allows alternative guidance laws to be substituted without modifying the integration layer. Descriptively higher agency and satisfaction ratings under haptic guidance support this interpretation. Planned system-level improvements include a stereoscopic VR display, user-adjustable force-feedback intensity, real-time motion scaling with singularity avoidance, and grip-integrated mode switching on the VerseGrip.}

\subsection{Limitations}
\label{sec:limitations}
\rev{Some limitations should be considered when interpreting these results. The hot-wire benchmark evaluates teleoperation primitives (path following, collision avoidance, speed regulation), not surgical skills; the results must therefore not be interpreted as validation of surgical performance. Visual feedback was monocular and 2D, unlike stereoscopic clinical consoles; the missing depth cues likely increased the reliance on haptic cues, so the measured benefit of guidance may be larger than in a stereoscopic setup. Conversely, the heterogeneous prior experience of the participants and the short, fixed training phase may have increased inter-subject variance and introduced within-session learning effects that counterbalancing cannot fully remove; in this respect, the between-mode differences are conservative estimates. Finally, since no subjective comparison survived Bonferroni correction, the subjective ratings constitute descriptive trends rather than confirmatory evidence. Future studies should employ homogeneous or covariate-adjusted samples, extended or criterion-based training, stereoscopic visualization, and surgically representative tasks.}

\section{Preliminary Integration on Physical da Vinci Hardware}
\label{sec:validation}
At the Antal Bejczy Center for Intelligent Robotics, the proposed haptic shared control framework was integrated with a real dVRK setup, enabling a preliminary demonstration in a physical teleoperation scenario.
The integration process involved interfacing the Haply Inverse3 and the VerseGrip with the dVRK control software, allowing for real-time haptic feedback during teleoperation tasks. 
Initial tests were conducted using a simplified teleoperation task in which the end-effector was guided along a wire trajectory.

\rev{During these initial tests, the same control-loop architecture as in the simulation study was used, and the system operated without instability or watchdog-triggered force interruption. Regarding quantitative performance:\\
(i)~\emph{Control-loop frequency:} although the Haply device supports a haptic refresh rate of up to \SI{4}{\kilo\hertz}, we matched the controller rate to the dVRK control loop at \SI{1}{\kilo\hertz}.\\
(ii)~\emph{Tracking accuracy:} the dVRK's positioning accuracy is in the range of \SIrange{1}{2}{\milli\meter} depending on hardware, tools, and workspace position, whereas the Inverse3 position resolution is \SI{0.01}{\milli\meter} according to the manufacturer; the tracking accuracy of the overall system is therefore dominated by the robot (slave) side.\\
(iii)~\emph{Communication delay:} without direct delay measurements, we can state that the integration did not noticeably increase the generally acceptable maximum latency of approximately \SI{200}{\milli\second} of the telemanipulation system.\\
(iv)~\emph{Stability:} the viscous damping component was included to suppress oscillatory behavior; during the physical experiments, no oscillatory instability or watchdog-triggered force interruption was observed.}

An initial video demonstration of the integrated system is available online\footnote{Video: \url{https://drive.google.com/file/d/1rhxtOhJYAiZa6y_rpFKxcEMbYJMQ-s_D/view}}, showcasing the real-time haptic feedback and control capabilities of the system. The video illustrates the operator manipulating the haptic device while receiving force feedback corresponding to interactions with the virtual environment, demonstrating the feasibility of the proposed haptic shared control framework in a real-world setting. In Figure~\ref{fig:dv_integration}, the demonstration setup is shown, highlighting the physical setup of the dVRK and the Haply Inverse3 haptic device during the teleoperation task.

\section{Conclusion and Outlook}
\label{sec:conclusion}
This paper presented a haptic shared-control framework for surgical teleoperation based on the integration of the Haply Inverse3 with the da~Vinci Research Kit. The system combines a ROS\,2-based software architecture with mapping strategies and virtual-fixture-inspired haptic guidance for a teleoperated hot-wire task. The user study showed that linear haptic guidance improved performance over manual control, yielding fewer wire contacts and shorter completion times. \rev{In contrast, the progressive-stiffness guidance produced consistent but statistically non-significant improvements in the same direction, suggesting that the potential benefits of more complex guidance profiles may not outweigh their added perceptual and cognitive demands, at least for novice operators.} The framework was also integrated with a physical dVRK setup, demonstrating technical feasibility, although quantitative hardware validation remains future work.

Future studies will examine cognitive load, task difficulty, and adaptive assistance strategies. Formal interaction models, such as ordinal potential differential games, may support adaptive shared control. Further work will focus on stereoscopic visualization, larger homogeneous user groups, quantitative comparison of simulated and physical setups, and cognitive-load-aware haptic guidance.

\section*{Acknowledgements}
This work is related to the MedLaBotX project (2024-1.2.3-HU-RIZONT-00069).
Project 2024-1.2.3-HU-RIZONT-00069 has been implemented with support provided by the Ministry of Culture and Innovation of Hungary from the National Research, Development, and Innovation Fund, financed under the 2024-1.2.3-HU-RIZONT funding scheme.


\begin{thebibliography}{44}
\providecommand{\url}[1]{#1}
\csname url@samestyle\endcsname
\providecommand{\newblock}{\relax}
\providecommand{\bibinfo}[2]{#2}
\providecommand{\BIBentrySTDinterwordspacing}{\spaceskip=0pt\relax}
\providecommand{\BIBentryALTinterwordstretchfactor}{4}
\providecommand{\BIBentryALTinterwordspacing}{\spaceskip=\fontdimen2\font plus
\BIBentryALTinterwordstretchfactor\fontdimen3\font minus
  \fontdimen4\font\relax}
\providecommand{\BIBforeignlanguage}[2]{{%
\expandafter\ifx\csname l@#1\endcsname\relax
\typeout{** WARNING: IEEEtran.bst: No hyphenation pattern has been}%
\typeout{** loaded for the language `#1'. Using the pattern for}%
\typeout{** the default language instead.}%
\else
\language=\csname l@#1\endcsname
\fi
#2}}
\providecommand{\BIBdecl}{\relax}
\BIBdecl

\bibitem{Okamura2009}
A.~M. Okamura, ``Haptic feedback in robot-assisted minimally invasive
  surgery,'' \emph{Current Opinion in Urology}, vol.~19, no.~1, pp. 102--107,
  2009.

\bibitem{Kitagawa2005SensorySubstitution}
\BIBentryALTinterwordspacing
M.~Kitagawa, D.~Dokko, A.~M. Okamura, and D.~D. Yuh, ``Effect of sensory
  substitution on suture-manipulation forces for robotic surgical systems,''
  \emph{The Journal of Thoracic and Cardiovascular Surgery}, vol. 129, no.~1,
  pp. 151--158, 2005. 
\BIBentrySTDinterwordspacing

\bibitem{Bergholz2023}
M.~Bergholz, M.~Ferle, and B.~M. Weber, ``The benefits of haptic feedback in
  robot assisted surgery and their moderators: A meta-analysis,''
  \emph{Scientific Reports}, vol.~13, p. 19215, 2023.

\bibitem{Patel2022}
R.~V. Patel, S.~F. Atashzar, and M.~Tavakoli, ``Haptic feedback and force-based
  teleoperation in surgical robotics,'' \emph{Proceedings of the IEEE}, vol.
  110, no.~7, pp. 1012--1027, 2022.

\bibitem{Awad2024}
M.~M. Awad, M.~C. Raynor, M.~Padmanabhan-Kabana, L.~Y. Schumacher, and
 J.~A. Blatnik, ``Evaluation of forces applied to tissues during
 robotic-assisted surgical tasks using a novel force feedback technology,''
 \emph{Surgical Endoscopy}, vol.~38, no.~10, pp. 6193--6202, 2024.

\bibitem{Servais2025}
E.~L. Servais, L.~Rashidi, P.~Porwal, M.~Garibaldi, and A.~J. Hung, ``Novel
  force feedback technology improves suturing in robotic-assisted surgery: A
  pre-clinical study,'' \emph{Surgical Endoscopy}, vol.~39, no.~2, pp.
  1217--1226, 2025.

\bibitem{Chang2025}
G.~C. Chang, U.~Sidhu, D.~Lai, G.~Vargas, S.~Shah, and L.~Rashidi, ``A pilot
  study on the impact of {da Vinci}~5's force feedback on clinical outcomes:
  {D}oes it lead to less tissue trauma in colorectal surgery?'' \emph{The
  American Journal of Surgery}, p. 116518, 2025.

\bibitem{Kneuertz2025}
P.~J. Kneuertz, R.~Mostellar, R.~E. Merritt, E.~L. Servais, B.~Mitzman, N.~R.
  Villamizar, L.~F. Tapias, J.~F. Lazar, D.~M. D'Souza, D.~S. Oh, and G.~P.
  Jackson, ``Force in robotic thoracic surgery---a one year analysis of
  {DaVinci}~5 force feedback,'' \emph{Journal of Robotic Surgery}, vol.~19,
  no.~1, p. 632, 2025.

\bibitem{Grobbel_SMC2023}
M.~Grobbel~et al., ``Shared telemanipulation with vr controllers in an anti
  slosh scenario,'' in \emph{2023 IEEE International Conference on Systems,
  Man, and Cybernetics (SMC)}, 2023, pp. 4405--4410.

\bibitem{VFlemisch_2024_AT}
\BIBentryALTinterwordspacing
B.~Varga, F.~Flemisch, and S.~Hohmann, ``Human in the loop,'' \emph{at -
  Automatisierungstechnik}, vol.~72, no.~12, pp. 1109--1111, 2024. 
\BIBentrySTDinterwordspacing

\bibitem{Varga2024LQSharedControl}
B.~Varga, ``Toward adaptive cooperation: Model-based shared control using
  {LQ}-differential games,'' \emph{Acta Polytechnica Hungarica}, vol.~21,
  no.~10, pp. 439--462, 2024.

\bibitem{Kazanzides2014}
P.~Kazanzides, Z.~Chen, A.~Deguet, G.~S. Fischer, R.~H. Taylor, and S.~P.
  DiMaio, ``An open-source research kit for the {da
  Vinci}\textsuperscript{\textregistered} {S}urgical {S}ystem,'' in \emph{Proc.
  IEEE International Conference on Robotics and Automation (ICRA)}, Hong Kong,
  China, 2014, pp. 6434--6439.

\bibitem{DEttorre2021}
C.~D'Ettorre, A.~Mariani, A.~Stilli, F.~Rodriguez~y Baena, P.~Valdastri,
  A.~Deguet, P.~Kazanzides, R.~H. Taylor, G.~S. Fischer, S.~P. DiMaio,
  A.~Menciassi, and D.~Stoyanov, ``Accelerating surgical robotics research: A
  review of 10 years with the {da Vinci Research Kit},'' \emph{IEEE Robotics \&
  Automation Magazine}, vol.~28, no.~4, pp. 56--78, 2021.

\bibitem{HaplyInverse3}
{Haply Robotics}, ``Inverse3 haptic controller --- product specification,''
  \url{https://www.haply.co/inverse3}, 2024, accessed 2026-06-02.

\bibitem{Okamura2004}
A.~M. Okamura, ``Methods for haptic feedback in teleoperated robot-assisted
  surgery,'' \emph{Industrial Robot: An International Journal}, vol.~31, no.~6,
  pp. 499--508, 2004.

\bibitem{hagen2008visual}
M.~Hagen, J.~Meehan, I.~Inan, and P.~Morel, ``Visual clues act as a substitute
  for haptic feedback in robotic surgery,'' \emph{Surgical endoscopy}, vol.~22,
  no.~6, pp. 1505--1508, 2008.

\bibitem{Enayati2016}
N.~Enayati, E.~De~Momi, and G.~Ferrigno, ``Haptics in robot-assisted surgery:
  Challenges and benefits,'' \emph{IEEE Reviews in Biomedical Engineering},
  vol.~9, pp. 49--65, 2016.

\bibitem{Abbink2012}
D.~A. Abbink, M.~Mulder, and E.~R. Boer, ``Haptic shared control: Smoothly
  shifting control authority?'' \emph{Cognition, Technology \& Work}, vol.~14,
  no.~1, pp. 19--28, 2012.

\bibitem{Abbink2018}
D.~A. Abbink, T.~Carlson, M.~Mulder, J.~C.~F. de~Winter, F.~Aminravan, T.~L.
  Gibo, and E.~R. Boer, ``A topology of shared control systems---finding common
  ground in diversity,'' \emph{IEEE Transactions on Human-Machine Systems},
  vol.~48, no.~5, pp. 509--525, 2018.

\bibitem{Boessenkool2013}
H.~Boessenkool, D.~A. Abbink, C.~J.~M. Heemskerk, F.~C.~T. van~der Helm, and
  J.~G.~W. Wildenbeest, ``A task-specific analysis of the benefit of haptic
  shared control during telemanipulation,'' \emph{IEEE Transactions on
  Haptics}, vol.~6, no.~1, pp. 2--12, 2013.

\bibitem{VanOosterhout2015}
J.~van Oosterhout, J.~G.~W. Wildenbeest, H.~Boessenkool, C.~J.~M. Heemskerk,
  M.~R. de~Baar, F.~C.~T. van~der Helm, and D.~A. Abbink, ``Haptic shared
  control in tele-manipulation: Effects of inaccuracies in guidance on task
  execution,'' \emph{IEEE Transactions on Haptics}, vol.~8, no.~2, pp.
  164--175, 2015.

\bibitem{Nudehi2005}
S.~S. Nudehi, R.~Mukherjee, and M.~Ghodoussi, ``A shared-control approach to
  haptic interface design for minimally invasive telesurgical training,''
  \emph{IEEE Transactions on Control Systems Technology}, vol.~13, no.~4, pp.
  588--592, 2005.

\bibitem{Selvaggio2018HSC}
M.~Selvaggio, F.~Abi-Farraj, C.~Pacchierotti, P.~Robuffo~Giordano, and
  B.~Siciliano, ``Haptic-based shared-control methods for a dual-arm system,''
  \emph{IEEE Robotics and Automation Letters}, vol.~3, no.~4, pp. 4249--4256,
  2018.

\bibitem{Selvaggio2021}
M.~Selvaggio, M.~Cognetti, S.~Nikolaidis, S.~Ivaldi, and B.~Siciliano,
  ``Autonomy in physical human-robot interaction: A brief survey,'' \emph{IEEE
  Robotics and Automation Letters}, vol.~6, no.~4, pp. 7989--7996, 2021.

\bibitem{Rosenberg1993}
L.~B. Rosenberg, ``Virtual fixtures: Perceptual tools for telerobotic
  manipulation,'' in \emph{Proc. IEEE Virtual Reality Annual International
  Symposium (VRAIS)}, Seattle, WA, USA, 1993, pp. 76--82.

\bibitem{2025_SharedControlApproach_varga}
B.~Varga and M.~Poncelet, ``A {{Shared Control Approach}} to {{Robot-Assisted
  Cataract Surgery Training}} for {{Novice Surgeons}},'' \emph{Sensors},
  vol.~25, no.~16, p. 5165, Aug. 2025.

\bibitem{Abbott2007}
J.~J. Abbott, P.~Marayong, and A.~M. Okamura, ``Haptic virtual fixtures for
  robot-assisted manipulation,'' in \emph{Robotics Research}, ser. Springer
  Tracts in Advanced Robotics, S.~Thrun, R.~Brooks, and H.~Durrant-Whyte,
  Eds.\hskip 1em plus 0.5em minus 0.4em\relax Berlin: Springer, 2007, vol.~28,
  pp. 49--64.

\bibitem{Bowyer2014}
S.~A. Bowyer, B.~L. Davies, and F.~Rodriguez~y Baena, ``Active
  constraints/virtual fixtures: A survey,'' \emph{IEEE Transactions on
  Robotics}, vol.~30, no.~1, pp. 138--157, 2014.

\bibitem{Ferraguti2015}
F.~Ferraguti, N.~Preda, A.~Manurung, M.~Bonf{\`e}, O.~Lambercy, R.~Gassert,
  R.~Muradore, P.~Fiorini, and C.~Secchi, ``An energy tank-based interactive
  control architecture for autonomous and teleoperated robotic surgery,''
  \emph{IEEE Transactions on Robotics}, vol.~31, no.~5, pp. 1073--1088, 2015.

\bibitem{Selvaggio2019ICRA}
M.~Selvaggio, P.~Robuffo~Giordano, F.~Ficuciello, and B.~Siciliano, ``Passive
  task-prioritized shared-control teleoperation with haptic guidance,'' in
  \emph{Proc. IEEE International Conference on Robotics and Automation (ICRA)},
  Montreal, Canada, 2019, pp. 430--436.

\bibitem{Selvaggio2018VF}
M.~Selvaggio, G.~A. Fontanelli, F.~Ficuciello, L.~Villani, and B.~Siciliano,
  ``Passive virtual fixtures adaptation in minimally invasive robotic
  surgery,'' \emph{IEEE Robotics and Automation Letters}, vol.~3, no.~4, pp.
  3129--3136, 2018.

\bibitem{Tobergte2011}
A.~Tobergte, P.~Helmer, U.~Hagn, P.~Rouiller, S.~Thielmann, S.~Grange,
  A.~Albu-Sch{\"a}ffer, F.~Conti, and G.~Hirzinger, ``The sigma.7 haptic
  interface for {MiroSurge}: A new bi-manual surgical console,'' in \emph{Proc.
  IEEE/RSJ International Conference on Intelligent Robots and Systems (IROS)},
  San Francisco, CA, USA, 2011, pp. 3023--3030.

\bibitem{Zareinia2015}
K.~Zareinia, Y.~Maddahi, C.~Ng, N.~Sepehri, and G.~R. Sutherland, ``Performance
  evaluation of haptic hand-controllers in a robot-assisted surgical system,''
  \emph{The International Journal of Medical Robotics and Computer Assisted
  Surgery}, vol.~11, no.~4, pp. 486--501, 2015.

\bibitem{Torabi2019}
A.~Torabi, M.~Khadem, K.~Zareinia, G.~R. Sutherland, and M.~Tavakoli,
  ``Application of a redundant haptic interface in enhancing soft-tissue
  stiffness discrimination,'' \emph{IEEE Robotics and Automation Letters},
  vol.~4, no.~2, pp. 1037--1044, 2019.

\bibitem{Pacchierotti2024}
C.~Pacchierotti and D.~Prattichizzo, ``Cutaneous/tactile haptic feedback in
  robotic teleoperation: Motivation, survey, and perspectives,'' \emph{IEEE
  Transactions on Robotics}, vol.~40, pp. 978--998, 2024.

\bibitem{10161348}
C.~Braun~et al., ``Using a collaborative robotic arm as human-machine
  interface: System setup and application to pose control tasks,'' in
  \emph{2023 IEEE International Conference on Robotics and Automation (ICRA)},
  2023, pp. 12\,464--12\,470.

\bibitem{Gomez2025}
E.~D. Gomez, H.~Mat~Husin, K.~R. Dumon, N.~N. Williams, and K.~J. Kuchenbecker,
  ``Simulation training with haptic feedback of instrument vibrations reduces
  resident workload during live robot-assisted sleeve gastrectomy,''
  \emph{Surgical Endoscopy}, vol.~39, pp. 1523--1535, 2025.

\bibitem{Yilmaz2024Enhancing}
N.~Yilmaz, B.~Burkhart, A.~Deguet, P.~Kazanzides, and U.~Tumerdem, ``Enhancing
  robotic telesurgery with sensorless haptic feedback,'' \emph{International
  Journal of Computer Assisted Radiology and Surgery}, vol.~19, pp. 1147--1155,
  2024.

\bibitem{Schleer2020}
P.~Schleer, P.~Kaiser, S.~Drobinsky, and K.~Radermacher, ``Augmentation of
  haptic feedback for teleoperated robotic surgery,'' \emph{International
  Journal of Computer Assisted Radiology and Surgery}, vol.~15, pp. 515--529,
  2020.

\bibitem{Abiri2019}
A.~Abiri, J.~Pensa, A.~Tao, J.~Ma, Y.-Y. Juo, S.~J. Askari, J.~Bisley,
  J.~Rosen, E.~P. Dutson, and W.~S. Grundfest, ``Multi-modal haptic feedback
  for grip force reduction in robotic surgery,'' \emph{Scientific Reports},
  vol.~9, p. 5016, 2019.

\bibitem{Quek2019}
Z.~F. Quek, W.~R. Provancher, and A.~M. Okamura, ``Evaluation of skin
  deformation tactile feedback for teleoperated surgical tasks,'' \emph{IEEE
  Transactions on Haptics}, vol.~12, no.~2, pp. 102--113, 2019.
\newpage 

\bibitem{Fontanelli2020}
G.~A. Fontanelli, L.~R. Buonocore, F.~Ficuciello, L.~Villani, and B.~Siciliano,
  ``An external force sensing system for minimally invasive robotic surgery,''
  \emph{IEEE/ASME Transactions on Mechatronics}, vol.~25, no.~3, pp.
  1543--1554, 2020.

\bibitem{Yilmaz2024Sensorless}
N.~Yilmaz, B.~Burkhart, A.~Deguet, P.~Kazanzides, and U.~Tumerdem, ``Sensorless
  transparency optimized haptic teleoperation on the {da Vinci Research Kit},''
  \emph{IEEE Robotics and Automation Letters}, vol.~9, no.~2, pp. 971--978,
  2024.

\bibitem{Saracino2019}
A.~Saracino, A.~Deguet, F.~Staderini, M.~N. Boushaki, F.~Cianchi, A.~Menciassi,
  and E.~Sinibaldi, ``Haptic feedback in the {da Vinci Research Kit} ({dVRK}):
  A user study based on grasping, palpation, and incision tasks,'' \emph{The
  International Journal of Medical Robotics and Computer Assisted Surgery},
  vol.~15, no.~4, p. e1999, 2019.
  

\end{thebibliography}

\end{document}